\documentclass[runningheads]{llncs}
\usepackage[T1]{fontenc}
\usepackage{graphicx}
\usepackage{amsmath}
\usepackage{amssymb}
\usepackage{booktabs}
\begin{document}
\title{Multi-granular Training Strategies for Robust Multi-hop Reasoning Over Noisy and Heterogeneous Knowledge Sources}
\titlerunning{AMKOR}
%
\author{Jackson Coleman,  Isaiah Lawrence, Benjamin Turner}
\authorrunning{J. Coleman et al.}
%
\institute{Universidad Autónoma de Asunción}
\maketitle              
\begin{abstract}
Multi-source multi-hop question answering (QA) represents a challenging task in natural language processing due to the need for dynamic integration of heterogeneous knowledge sources and multi-step reasoning. Existing methods often suffer from cascading errors, insufficient handling of knowledge conflicts, and computational inefficiency. In this paper, we propose Adaptive Multi-source Knowledge-Oriented Reasoning (AMKOR), a generative framework that leverages large language models (LLMs) to dynamically fuse parametric and retrieved knowledge while exploring reasoning trajectories using probabilistic beam reasoning. AMKOR is further enhanced by a multi-granular learning strategy, optimizing both local reasoning steps and global answer accuracy. Experiments conducted on four widely-used multi-hop QA datasets, including HotpotQA and MuSiQue, demonstrate that AMKOR achieves state-of-the-art performance, significantly outperforming baseline methods on both reasoning accuracy and robustness. Additional analyses confirm its scalability, adaptability to noisy knowledge, and superior ability to handle complex multi-hop tasks. This work establishes a new benchmark for multi-source multi-hop QA by effectively combining reasoning quality and efficiency.
\end{abstract}

\section{Introduction}

Multi-source multi-hop question answering (QA) has emerged as a critical area in natural language processing (NLP), given its significance in knowledge-intensive tasks such as open-domain reasoning, knowledge retrieval, and decision support \cite{zhou2023towards,zhou2024fine}. Unlike single-hop QA, which requires retrieving or reasoning over one piece of information, multi-hop QA involves connecting multiple reasoning steps, often over diverse sources of knowledge. This requirement makes it essential for systems to dynamically integrate information from heterogeneous sources such as structured databases, unstructured text, and even parametric knowledge stored in large language models (LLMs). Success in this domain not only advances the state of the art in QA but also contributes to the development of interpretable and robust AI systems capable of solving complex real-world problems~\cite{Wei2022Chain,zhou2021modeling}. 

However, multi-source multi-hop QA is inherently challenging due to three key factors: (1) \textbf{Knowledge diversity and conflicts}: Information from different sources often overlaps or contradicts, making it difficult for models to aggregate and reason over heterogeneous knowledge~\cite{TTQA_RS}. (2) \textbf{Error propagation}: In multi-hop settings, intermediate reasoning steps heavily influence the final answer, leading to cascading errors that amplify inaccuracies~\cite{ContrastiveCoT}. (3) \textbf{Scalability}: Existing methods typically rely on iterative retrieval or exhaustive reasoning over all possible paths, which is computationally expensive and impractical for real-world deployment. These challenges highlight the need for approaches that are not only accurate but also efficient and capable of resolving knowledge conflicts dynamically during reasoning.

Our work is motivated by recent advancements in LLMs, which have demonstrated strong reasoning capabilities with techniques like chain-of-thought prompting~\cite{Wei2022Chain}. While LLMs excel at leveraging implicit knowledge stored within their parameters, they are limited when faced with questions requiring external information retrieval, particularly in cases involving multi-hop reasoning. Inspired by these observations, we propose a novel framework, \textbf{Adaptive Multi-source Knowledge-Oriented Reasoning (AMKOR)}, which combines the generalization ability of LLMs with a robust knowledge fusion mechanism. AMKOR is designed to dynamically aggregate multi-source knowledge, reason over intermediate steps effectively, and minimize cascading errors. This is achieved through end-to-end training of the LLM integrated with a knowledge fusion module that uses probabilistic reasoning and contrastive learning to align retrieved and parametric knowledge~\cite{ProQA2022}.

We evaluate our approach on four widely-used open-domain multi-hop QA datasets: \textbf{HotpotQA}, \textbf{2WikiMQA}, \textbf{MuSiQue}, and \textbf{Bamboogle}. The datasets represent a diverse range of multi-hop reasoning tasks, including bridge questions, comparison tasks, and real-world complex queries~\cite{TTQA_RS,ProQA2022}. The performance is measured using token-level F1 scores to ensure precise evaluation of the model's reasoning and answer accuracy. Our experimental results show that AMKOR outperforms state-of-the-art methods, achieving an average improvement of 2.5\% across all datasets. Notably, AMKOR demonstrates superior performance on datasets with higher reasoning complexity, such as MuSiQue and Bamboogle, showcasing its ability to handle challenging multi-hop reasoning scenarios effectively.

In summary, the contributions of our work are as follows:
\begin{itemize}
    \item We propose \textbf{AMKOR}, a novel multi-hop QA framework that dynamically integrates multi-source knowledge and mitigates cascading errors through probabilistic reasoning and contrastive learning.
    \item We develop an end-to-end training strategy with multi-granular loss functions, enabling the model to optimize both local (sub-question) and global (overall reasoning) consistency.
    \item We demonstrate the effectiveness of AMKOR on four multi-hop QA datasets, achieving state-of-the-art performance and establishing its scalability to both structured and unstructured data sources.
\end{itemize}

\section{Related Work}

\subsection{Large Language Models}
Deep learning has achieved remarkable results in computer vision \cite{wang2024insectmamba} and natural language processing, such as large language models.
Large language models (LLMs) have become a cornerstone of modern natural language processing, showcasing remarkable capabilities in various tasks such as text generation, translation, and reasoning \cite{zhou2022claret}. The scalability of these models, achieved by increasing the number of parameters and training on massive datasets, has enabled them to generalize across diverse domains and tasks \cite{nicholas2023lost,zhou2024rethinking,muller2022cedille,zhou2025training}.

Recent studies have explored the multilingual capabilities of LLMs, addressing the challenges faced by non-English and low-resource languages. While multilingual models have demonstrated strong generalization across languages, their performance often lags behind monolingual models for specific tasks, particularly for low-resource languages \cite{chang2024goldfish,ojo2023african}. To address these limitations, domain-specific models have been developed, such as Cedille, which focuses on the French language and outperforms multilingual counterparts on French-specific benchmarks \cite{muller2022cedille}.

LLMs have also shown promise beyond conventional NLP tasks. For example, they have been applied to specialized domains such as bioinformatics and medical image registration, leveraging their ability to encode complex patterns from structured and unstructured data \cite{liu2024bioinformatics,ma2024llama}. These applications highlight the versatility of LLMs in solving domain-specific challenges and advancing fields beyond language processing.

Despite their success, LLMs face criticism regarding their suitability as comprehensive models of human linguistic understanding. Some argue that LLMs excel at modeling language but fall short of representing cognitive or social aspects of human language \cite{veres2022precis,grindrod2024modelling}. This has sparked discussions on the limitations and potential biases of LLMs, particularly in psycholinguistics and their use as tools for scientific inquiry \cite{houghton2023psycholinguistics}.
Large language models demonstrate strong scalability in in-context leanring \cite{zhou2024visual}, and existing work has also demonstrated that they can be improved through the guidance of weak models \cite{zhou2025weak}.

\subsection{Multi-hop Question Answering}

Multi-hop question answering (QA) tasks require a model to retrieve and integrate information from multiple sources to answer complex questions. This type of reasoning often involves multiple steps and connections across diverse pieces of evidence, making it a critical area of research in natural language processing. Recent advances in this field have introduced methods that address various challenges, such as improving retrieval accuracy, handling diverse answer types, and ensuring reasoning consistency \cite{balepur2024reverse,amouyal2022qampari}.

A significant challenge in multi-hop QA lies in the ability of models to decompose complex questions into intermediate sub-questions, a skill that mirrors human reasoning. Research has shown that many state-of-the-art systems can answer multi-hop questions without fully understanding the intermediate reasoning paths, relying instead on partial clues from the dataset \cite{tang2020subquestions}. Methods that explicitly generate and answer sub-questions have demonstrated improved interpretability and robustness in these tasks \cite{wang2022covqa}.

Another line of work has focused on developing benchmarks and datasets to evaluate multi-hop QA systems. These benchmarks, such as QAMPARI, are specifically designed to include questions that require reasoning over multiple paragraphs or sources, providing a more realistic assessment of multi-hop capabilities \cite{amouyal2022qampari}. Similarly, work on dataset generation has explored methods to automatically create conversational multi-hop QA datasets with revised answers for consistency \cite{hwang2022conversational}.

Recent approaches have also explored the use of advanced neural architectures and knowledge-enhanced mechanisms. For instance, incorporating context information such as entity types and relational connections has been shown to improve the selection of relevant evidence and enhance model performance on simple and multi-hop questions \cite{chao2018context}. Furthermore, multi-hop QA has been applied to domains such as visual question answering and community question answering, highlighting its versatility and expanding its application scope \cite{hu2023qan,zhang2020product}.

In summary, the field of multi-hop QA has seen substantial progress, with efforts focused on improving reasoning mechanisms, developing robust evaluation benchmarks, and exploring applications in diverse domains. However, challenges such as reasoning consistency, scalability, and handling noisy or incomplete evidence remain open areas for further exploration.

\section{Method}

In this section, we introduce our proposed generative framework, \textbf{Adaptive Multi-source Knowledge-Oriented Reasoning (AMKOR)}, designed to tackle multi-source multi-hop question answering (QA). AMKOR integrates parametric knowledge stored in large language models (LLMs) with external knowledge retrieved from heterogeneous sources. The framework leverages dynamic multi-source knowledge fusion, probabilistic reasoning, and a tailored multi-granular learning strategy to address the challenges of knowledge conflicts, cascading errors, and scalability. Below, we detail the components of the method and the associated training strategy.

\subsection{Model Overview}

Let the input query be represented as \( q \), and the available knowledge sources be \( \mathcal{K}_1, \mathcal{K}_2, \dots, \mathcal{K}_n \). Our goal is to generate the final answer \( a \) by reasoning through a trajectory of intermediate steps \( \mathcal{T} = \{t_1, t_2, \dots, t_m\} \), where each step \( t_i \) corresponds to a specific reasoning operation. The overall model is generative in nature and estimates the joint probability of the reasoning trajectory and the final answer as:
\begin{align}
P(a, \mathcal{T} | q, \mathcal{K}) = P(a | \mathcal{T}, q, \mathcal{K}) \prod_{i=1}^m P(t_i | t_{<i}, q, \mathcal{K}),
\end{align}
where \( t_{<i} \) denotes all reasoning steps prior to \( t_i \), and \( \mathcal{K} \) represents the fused knowledge from all sources.

AMKOR comprises three main components: (1) multi-source knowledge retrieval and fusion, (2) probabilistic beam reasoning for trajectory exploration, and (3) a multi-granular loss function for effective learning.

\subsection{Multi-source Knowledge Retrieval and Fusion}

To address the issue of knowledge conflicts and omissions, AMKOR retrieves relevant information from both parametric knowledge (\( \mathcal{K}_{\text{param}} \)) and external sources (\( \mathcal{K}_{\text{ext}} \)) such as Wikipedia or web search. At each reasoning step \( t_i \), the retrieved knowledge snippets \( \{k_1, k_2, \dots, k_s\} \) are dynamically fused into a single representation. 

The knowledge fusion process is modeled as:
\begin{align}
\mathbf{h}_i = \text{Fuse}(\mathbf{h}_{\text{param}}, \{\mathbf{h}_k\}_{k=1}^s),
\end{align}
where \( \mathbf{h}_{\text{param}} \) is the embedding of parametric knowledge from the LLM, \( \{\mathbf{h}_k\}_{k=1}^s \) are embeddings of the retrieved snippets, and \( \text{Fuse}(\cdot) \) represents the fusion function.

We implement \( \text{Fuse}(\cdot) \) using a scaled dot-product attention mechanism:
\begin{align}
\text{Fuse}(\mathbf{h}_{\text{param}}, \{\mathbf{h}_k\}_{k=1}^s) = \text{Softmax} \left( \frac{\mathbf{Q} \mathbf{K}^\top}{\sqrt{d}} \right) \mathbf{V},
\end{align}
where \( \mathbf{Q}, \mathbf{K}, \mathbf{V} \) are query, key, and value matrices derived from \( \mathbf{h}_{\text{param}} \) and \( \{\mathbf{h}_k\} \), and \( d \) is the dimensionality of the embeddings.

The fused knowledge representation \( \mathbf{h}_i \) is then used to guide reasoning for the next step.

\subsection{Probabilistic Beam Reasoning}

To explore multiple reasoning trajectories, AMKOR employs a probabilistic beam reasoning strategy. At each reasoning step \( t_i \), the model generates a set of candidate steps \( \{t_i^1, t_i^2, \dots, t_i^b\} \) with associated probabilities:
\begin{align}
P(t_i^j | t_{<i}, q, \mathcal{K}) = \text{Softmax}(\mathbf{W} \mathbf{h}_i^j + \mathbf{b}),
\end{align}
where \( \mathbf{W} \) and \( \mathbf{b} \) are trainable parameters, and \( \mathbf{h}_i^j \) is the knowledge representation for candidate \( j \) at step \( i \).

The optimal reasoning trajectory \( \mathcal{T}^* \) is selected by maximizing the joint probability over all reasoning steps:
\begin{align}
\mathcal{T}^* = \arg\max_{\mathcal{T}} \prod_{i=1}^m P(t_i | t_{<i}, q, \mathcal{K}).
\end{align}

\subsection{Learning Strategy}

We propose a multi-granular learning strategy to train the model effectively. The total loss \( \mathcal{L} \) is designed to optimize both local reasoning steps and the global answer generation. It is defined as:
\begin{align}
\mathcal{L} = \lambda_{\text{local}} \mathcal{L}_{\text{local}} + \lambda_{\text{global}} \mathcal{L}_{\text{global}},
\end{align}
where \( \lambda_{\text{local}} \) and \( \lambda_{\text{global}} \) are hyperparameters balancing the contributions of local and global losses.

\paragraph{Local Loss:}
The local loss ensures accurate generation of intermediate reasoning steps and is given by:
\begin{align}
\mathcal{L}_{\text{local}} = -\frac{1}{m} \sum_{i=1}^m \log P(t_i | t_{<i}, q, \mathcal{K}).
\end{align}

\paragraph{Global Loss:}
The global loss evaluates the consistency of the final answer with the ground truth:
\begin{align}
\mathcal{L}_{\text{global}} = -\log P(a | \mathcal{T}, q, \mathcal{K}),
\end{align}
where \( a \) is the generated answer, and \( \mathcal{T} \) is the reasoning trajectory.

\subsection{Training and Inference}

During training, the model is optimized to maximize the likelihood of generating both accurate reasoning steps and correct final answers. The learning strategy uses stochastic gradient descent over the combined loss \( \mathcal{L} \). 

At inference, beam search is employed to explore multiple reasoning trajectories efficiently. By dynamically fusing knowledge and prioritizing high-probability paths, AMKOR achieves robust performance on multi-source multi-hop QA tasks while minimizing computational overhead.

\section{Experiments}

In this section, we evaluate the performance of our proposed framework, \textbf{Adaptive Multi-source Knowledge-Oriented Reasoning (AMKOR)}, through extensive experiments. We compare AMKOR with multiple state-of-the-art methods on widely-used multi-hop QA datasets. Additionally, we conduct ablation studies to validate the effectiveness of individual components and a human evaluation to further assess the quality of generated answers.

\subsection{Experimental Setup}

We conduct experiments on four benchmark multi-hop QA datasets:
\begin{itemize}
    \item \textbf{HotpotQA}: A dataset focused on two-hop reasoning tasks involving bridge and comparison questions.
    \item \textbf{2WikiMQA}: A dataset requiring multi-source reasoning over semi-structured and unstructured knowledge.
    \item \textbf{MuSiQue}: A challenging dataset involving complex multi-hop questions with 2-4 reasoning hops.
    \item \textbf{Bamboogle}: A dataset for real-world multi-hop QA tasks, requiring robust knowledge integration.
\end{itemize}

We use token-level F1 scores as the primary evaluation metric. The methods we compare include:
\begin{itemize}
    \item \textbf{Chain-of-Thought Prompting (CoT)}: A generative reasoning approach with intermediate reasoning steps.
    \item \textbf{One-time Retrieval (OneR)}: A single-retrieval-based QA framework.
    \item \textbf{IRCoT}: An iterative retrieval and reasoning framework.
    \item \textbf{FLARE}: A dynamic retrieval timing framework based on reasoning confidence.
    \item \textbf{ProbTree}: A tree-based reasoning model for probabilistic aggregation of answers.
\end{itemize}

\subsection{Performance Comparison}

Table~\ref{tab:main_results} presents the performance comparison between AMKOR and baseline methods. Our method achieves state-of-the-art results on all datasets, with substantial improvements in datasets requiring complex multi-hop reasoning, such as MuSiQue and Bamboogle. The results demonstrate AMKOR's ability to dynamically integrate multi-source knowledge and effectively explore reasoning trajectories.

\begin{table}[ht]
\centering
\caption{Performance comparison on multi-hop QA datasets (F1 score). The best results are highlighted in \textbf{bold}.}
\label{tab:main_results}
\begin{tabular}{lcccc}
\toprule
\textbf{Method} & \textbf{HotpotQA} & \textbf{2WikiMQA} & \textbf{MuSiQue} & \textbf{Bamboogle} \\
\midrule
Chain-of-Thought (CoT) & 46.5 & 42.3 & 24.7 & 53.6 \\
One-time Retrieval (OneR) & 55.3 & 42.9 & 16.4 & 46.8 \\
IRCoT & 56.2 & 56.8 & 24.9 & 55.0 \\
FLARE & 56.1 & 60.1 & 31.9 & 58.1 \\
ProbTree & 60.4 & 67.9 & 32.9 & 66.6 \\
\textbf{AMKOR (ours)} & \textbf{63.2} & \textbf{73.4} & \textbf{37.4} & \textbf{75.2} \\
\bottomrule
\end{tabular}
\end{table}

\subsection{Ablation Study}

To validate the effectiveness of AMKOR's individual components, we conduct an ablation study on the MuSiQue dataset. We evaluate the following settings:
\begin{itemize}
    \item \textbf{w/o Multi-source Fusion}: Removing the fusion of parametric and retrieved knowledge.
    \item \textbf{w/o Probabilistic Beam Reasoning}: Replacing probabilistic reasoning with greedy selection.
    \item \textbf{w/o Multi-granular Loss}: Removing the multi-granular loss function and using only a global loss.
\end{itemize}

Table~\ref{tab:ablation} shows the results of the ablation study. Removing probabilistic beam reasoning causes the largest performance drop, indicating its critical role in reducing cascading errors and ensuring effective reasoning over complex trajectories.

\begin{table}[ht]
\centering
\caption{Ablation study on the MuSiQue dataset (F1 score).}
\label{tab:ablation}
\begin{tabular}{lc}
\toprule
\textbf{Setting} & \textbf{F1 Score} \\
\midrule
Full AMKOR & \textbf{37.4} \\
w/o Multi-source Fusion & 33.1 \\
w/o Probabilistic Beam Reasoning & 30.9 \\
w/o Multi-granular Loss & 32.8 \\
\bottomrule
\end{tabular}
\end{table}

\subsection{Human Evaluation}

We further conduct a human evaluation to assess the quality of generated answers in terms of correctness and interpretability. For this evaluation, we randomly select 100 examples from the Bamboogle dataset. Three human evaluators independently score the answers generated by each method on a scale of 1 (poor) to 5 (excellent). The final scores are averaged across the evaluators.

Table~\ref{tab:human_eval} summarizes the results. AMKOR achieves the highest scores in both correctness and interpretability, showcasing its ability to produce high-quality, human-like answers.

\begin{table}[ht]
\centering
\caption{Human evaluation results on the Bamboogle dataset (average score).}
\label{tab:human_eval}
\begin{tabular}{lcc}
\toprule
\textbf{Method} & \textbf{Correctness} & \textbf{Interpretability} \\
\midrule
Chain-of-Thought (CoT) & 3.5 & 3.8 \\
One-time Retrieval (OneR) & 3.2 & 3.5 \\
IRCoT & 4.0 & 4.1 \\
FLARE & 4.2 & 4.3 \\
ProbTree & 4.3 & 4.4 \\
\textbf{AMKOR (ours)} & \textbf{4.7} & \textbf{4.8} \\
\bottomrule
\end{tabular}
\end{table}

\subsection{Performance Across Complexity Levels}

To investigate AMKOR's ability to handle reasoning tasks of varying complexity, we analyze its performance on subsets of the MuSiQue dataset, categorized by reasoning depth (2-hop, 3-hop, and 4-hop questions). Table~\ref{tab:complexity_analysis} summarizes the results. The performance gap between AMKOR and baseline methods widens as the reasoning depth increases, demonstrating its robustness in handling complex multi-hop tasks. This improvement can be attributed to the probabilistic beam reasoning mechanism, which effectively explores and aggregates reasoning trajectories.

\begin{table}[ht]
\centering
\caption{Performance comparison on MuSiQue subsets categorized by reasoning depth (F1 score).}
\label{tab:complexity_analysis}
\begin{tabular}{lccc}
\toprule
\textbf{Method} & \textbf{2-hop} & \textbf{3-hop} & \textbf{4-hop} \\
\midrule
Chain-of-Thought (CoT) & 31.5 & 22.4 & 14.3 \\
One-time Retrieval (OneR) & 21.8 & 15.7 & 10.1 \\
ProbTree & 37.2 & 30.9 & 25.6 \\
\textbf{AMKOR (ours)} & \textbf{41.6} & \textbf{35.2} & \textbf{30.1} \\
\bottomrule
\end{tabular}
\end{table}

The results indicate that AMKOR is particularly effective in mitigating cascading errors that often arise in multi-hop reasoning. The probabilistic aggregation mechanism enables the model to explore multiple reasoning trajectories, reducing the risk of selecting incorrect intermediate steps.

\subsection{Efficiency Analysis}

We compare the efficiency of AMKOR with baseline methods in terms of the average number of retrievals per question and inference latency. Table~\ref{tab:efficiency} presents the results. Despite exploring multiple reasoning trajectories through probabilistic beam reasoning, AMKOR achieves competitive efficiency due to its ability to dynamically narrow down relevant retrievals and reduce redundant computations.

\begin{table}[ht]
\centering
\caption{Efficiency comparison: average retrievals and inference latency per question.}
\label{tab:efficiency}
\begin{tabular}{lcc}
\toprule
\textbf{Method} & \textbf{Avg. Retrievals} & \textbf{Latency (s)} \\
\midrule
Chain-of-Thought (CoT) & 1.0 & 0.9 \\
One-time Retrieval (OneR) & 1.0 & 0.8 \\
ProbTree & 2.5 & 1.5 \\
\textbf{AMKOR (ours)} & \textbf{2.3} & \textbf{1.2} \\
\bottomrule
\end{tabular}
\end{table}

The results highlight that AMKOR balances reasoning quality and efficiency by prioritizing high-relevance knowledge sources and pruning suboptimal reasoning paths during beam search.

\subsection{Error Analysis}

To understand the limitations of AMKOR, we perform an error analysis on the incorrect predictions from the MuSiQue and Bamboogle datasets. Errors are categorized into three types: knowledge omission, knowledge conflicts, and reasoning inaccuracies. The breakdown is shown in Table~\ref{tab:error_analysis}.

\begin{table}[ht]
\centering
\caption{Error analysis on MuSiQue and Bamboogle datasets.}
\label{tab:error_analysis}
\begin{tabular}{lccc}
\toprule
\textbf{Error Type} & \textbf{MuSiQue (\%)} & \textbf{Bamboogle (\%)} \\
\midrule
Knowledge Omission & 41.2 & 37.5 \\
Knowledge Conflicts & 35.4 & 32.8 \\
Reasoning Inaccuracies & 23.4 & 29.7 \\
\bottomrule
\end{tabular}
\end{table}

The analysis reveals that most errors arise from knowledge omission and conflicts, particularly in tasks requiring extensive external retrieval. These findings suggest that further improvements in retrieval mechanisms and fusion strategies could enhance AMKOR's performance.

\subsection{Robustness to Noisy Knowledge}

We assess AMKOR's robustness to noisy retrieval by injecting synthetic noise into the retrieved knowledge. Table~\ref{tab:noise_robustness} presents the F1 scores on the HotpotQA dataset with varying levels of noise. AMKOR demonstrates a smaller performance degradation compared to other methods, indicating its ability to handle noisy or conflicting information through its probabilistic reasoning mechanism.

\begin{table}[ht]
\centering
\caption{Robustness to noisy knowledge on HotpotQA (F1 score). Noise level refers to the percentage of irrelevant retrieved knowledge.}
\label{tab:noise_robustness}
\begin{tabular}{lccc}
\toprule
\textbf{Method} & \textbf{0\% Noise} & \textbf{20\% Noise} & \textbf{40\% Noise} \\
\midrule
Chain-of-Thought (CoT) & 46.5 & 39.2 & 32.1 \\
ProbTree & 60.4 & 51.7 & 43.8 \\
\textbf{AMKOR (ours)} & \textbf{63.2} & \textbf{58.9} & \textbf{54.3} \\
\bottomrule
\end{tabular}
\end{table}

\section{Conclusion}
In this paper, we introduced \textbf{Adaptive Multi-source Knowledge-Oriented Reasoning (AMKOR)}, a novel framework for multi-source multi-hop question answering. AMKOR addresses key challenges in this domain by integrating parametric and retrieved knowledge dynamically and employing probabilistic beam reasoning to explore diverse reasoning trajectories. Through a multi-granular learning strategy, the framework effectively balances local reasoning accuracy with global consistency, mitigating cascading errors and improving overall performance.

Extensive experiments on four multi-hop QA datasets demonstrated that AMKOR consistently outperforms state-of-the-art methods, particularly on tasks requiring deep multi-hop reasoning. Our analyses further highlight the model's robustness to noisy retrieval and its adaptability to questions of varying complexity. While AMKOR has shown significant advancements, challenges such as further reducing knowledge conflicts and handling extremely noisy environments remain open areas for future work. Overall, AMKOR represents a step forward in combining reasoning quality, efficiency, and interpretability for multi-hop QA tasks.
\bibliographystyle{splncs04}
\bibliography{mybibliography}
\end{document}